\documentclass{llncs}

\usepackage{amsmath}
\usepackage{amssymb}
\usepackage{amsfonts}

\usepackage{bbm}
\usepackage{enumerate}

\usepackage{color}
\usepackage{cite}
\usepackage{graphicx}
\usepackage{pstricks}
\usepackage{amssymb}
\usepackage{amsfonts}
\usepackage{amsmath}
\usepackage[amsmath,thmmarks]{ntheorem}
\usepackage{todonotes}
\usepackage{url}

\begin{document}

\title{Crowd disagreement about medical images is informative}

\author{Veronika Cheplygina$^1$, Josien P. W. Pluim$^{1,2}$}
\institute{Medical Image Analysis, Department of Biomedical Engineering,\\ Eindhoven University of Technology, The Netherlands
\and Image Sciences Institute, University Medical Center Utrecht, The Netherlands 
}

\maketitle

\begin{abstract}
Classifiers for medical image analysis are often trained with a single consensus label, based on combining labels given by experts or crowds. However, disagreement between annotators may be informative, and thus removing it may not be the best strategy. As a proof of concept, we predict whether a skin lesion from the ISIC 2017 dataset is a melanoma or not, based on crowd annotations of visual characteristics of that lesion. We compare using the mean annotations, illustrating consensus, to standard deviations and other distribution moments, illustrating disagreement. We show that the mean annotations perform best, but that the disagreement measures are still informative. We also make the crowd annotations used in this paper available at \url{https://figshare.com/s/5cbbce14647b66286544}.
\end{abstract}

\section{Introduction}

In medical image analysis, machine learning is increasingly used for addressing different tasks. These include segmentation (labeling pixels as belonging to different classes, such as organs), detection (localizing structures of interest, such as tumors) and diagnosis (labeling an entire scan as having a disease or not). Classifiers for these tasks are typically trained with ground truth - accepted labels for existing data. 

Often, labels for these tasks are decided by one or more experts who visually inspect the image. If multiple experts are available, their labels can be combined by a majority vote or another procedure, and the consensus is used for training the classifier. For example, in a study or predicting malignancy of lung nodules, ~\cite{hussein2017risk} average the malignancy scores provided by several experts. Similarly, in studies of crowdsourcing for labeling medical images, the crowd labels are often combined, for example using majority vote~\cite{oneil2017crowdsourcing}, median combining~\cite{cheplygina2016early} or clustering~\cite{maier2015crowdtruth}.

However, disagreement in the individual labels could be informative for classification, and training on a consensus label may not be the optimal strategy. For example, \cite{guan2017who} show that learning a weight for each expert when grading diabetic retinopathy is better than averaging the labels in advance. Similarly, \cite{dumitrache2018crowdsourcing} show that modeling ambiguity is informative when extracting term relationships from medical texts.

In this paper we study whether disagreement is informative more directly. We use crowd annotations, describing visual features of skin lesion images, as inputs to predict an expert label (diagnosis) as an output. Although removing disagreement leads to the best performances, we show that disagreement alone leads to better-than-random results. Therefore, the disagreement of these crowd labels could be an advantage when training a skin lesion classifier with these crowd labels as additional outputs. 

\section{Methods}

\subsection{Data}

\begin{figure}
   \centering
    \includegraphics[width=5cm]{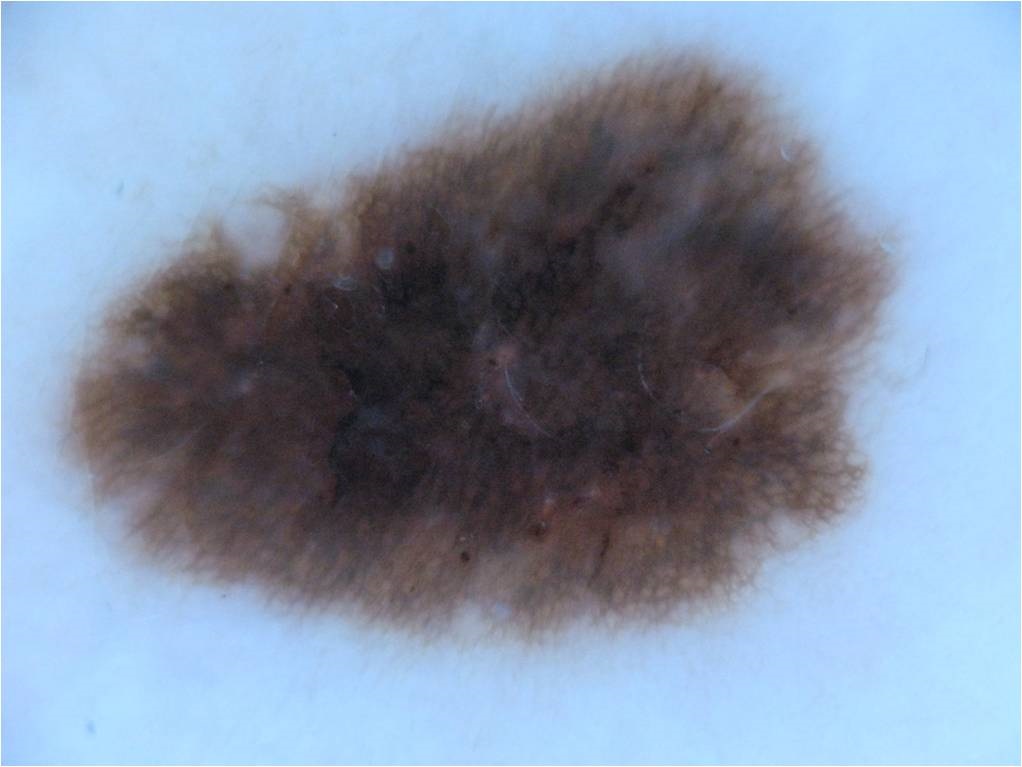}
     \includegraphics[width=5cm]{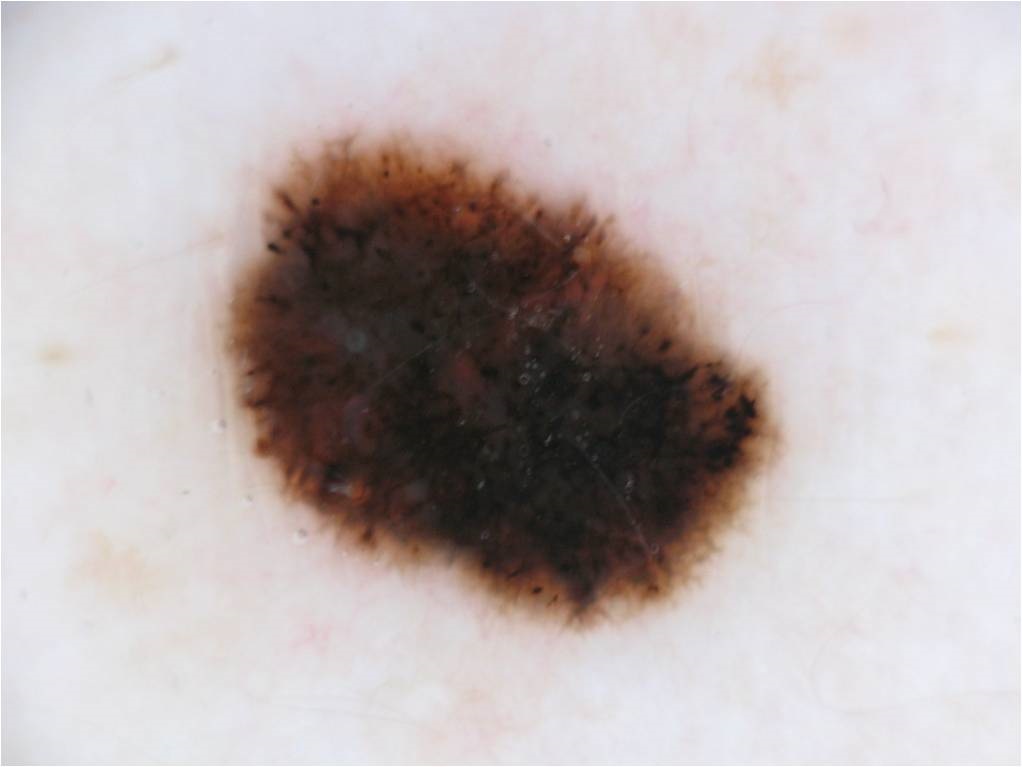}
    \caption{Examples of non-melanoma (left) and melanoma (right) images from the ISIC 2017 challenge}
    \label{fig:isic}    
\end{figure}

We collected the annotations during a first year undergraduate project course on medical image analysis (course code 8QA01, 2017-2018) at the Department of Biomedical Engineering, Eindhoven University of Technology. In groups of five or six, the students learned to automatically measure image features, such as ``asymmetry'', in images of skin lesions from the ISIC 2017 challenge~\cite{codella2017skin}, where one of the goals is to classify a lesion as melanoma or not. Examples of the images are shown in Fig.~\ref{fig:isic}.

The students also assessed such features visually, to be able to compare their algorithms' outputs to their own judgments. Each group was provided with a different set of 100 images of skin lesions, with approximately 20\% melanoma images. Each group was encouraged to decide which features they wanted to measure, invent their own way of grading the images, and assess each feature visually by at least three people. The students were not blinded to the melanoma/non-melanoma labels in the data, since the data is openly available online. 

An overview of the visual assessments collected is provided in Table \ref{tab:groups}. All groups annotated the ``ABC'' features - Asymmetry, Border and Color, some of the common features used by experts~\cite{abbasi2004early}. Some groups also annotated additional features such as the presence of dermoscopic structures, or added variations of the ABC features, for a total of eight different feature types. 

\begin{table}
\centering
\begin{tabular}{ p{1.5cm} p{2cm} p{2cm} l }

Group & Images/ annotator & Annotators/ feature & Features \\
  \hline			
   1 & 100 & 3 & A, B, C \\
   2 & 100 & 3 & A, B, C, C2\\
   3 & 100 & 3 & A, B, C\\
   4 & 100 & 3 & A, B, C, D\\
   5 & 50 & 3 & A, B, C, D, blood\\
   6 & 100 & 3 & A, B, C\\
   7 & 100 & 6 & A, B, C, D, G\\
   8 & 50 & 6 & A, B, C, B2 \\
\end{tabular}
\caption{Overview of features visually assessed by students: \textbf{A}symmetry, \textbf{B}order, \textbf{C}olor, \textbf{D}ermoscopic structures, blue \textbf{G}low. The number 2 indicates different variation}
\label{tab:groups}
\end{table}

In this paper we focus on one of the eight datasets collected by the groups,``group 7''. This group annotated a total of five different feature types: asymmetry of the lesion (scale 0-2), irregularity of the border (scale 0-2), number of colors present (scale 1-6), presence of structures such as dots (scale 0-2) and presence of a blueish glow (scale 0-2). Each feature type was annotated by six annotators per image, leading to 30 features in total. We removed four images with missing values, and normalized each feature to zero mean and unit variance before proceeding with the experiments. 

\begin{figure}
   \centering
    \includegraphics[width=0.5\textwidth]{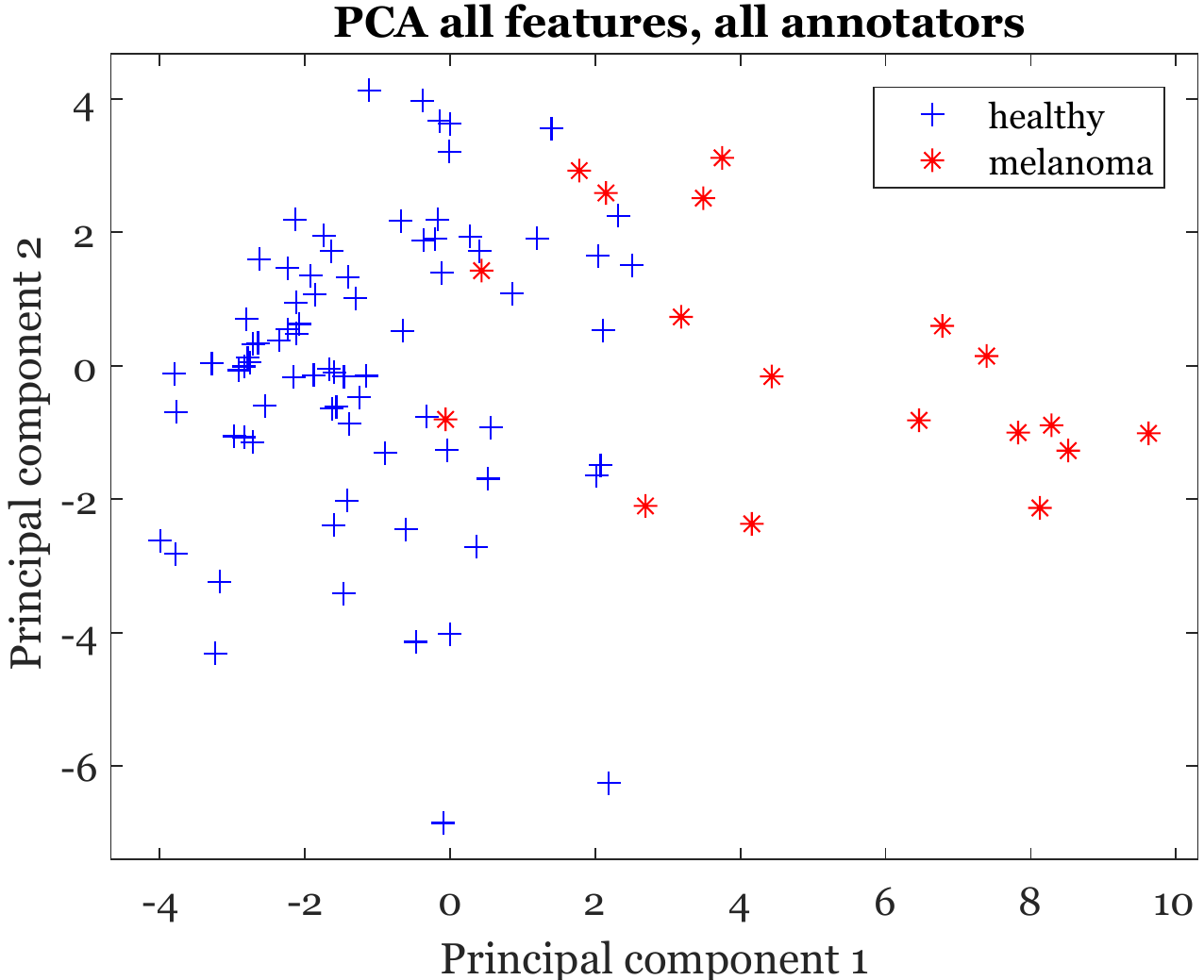}
    \caption{Principal component analysis embedding of 30-feature dataset of annotations}
    \label{fig:pca}
\end{figure}

An embedding of the first two principal components of the 30 dimensional dataset is shown in Fig.~\ref{fig:pca}. This plot indicates that the visual attributes provided by the group already provide a good separation between the melanoma and non-melanoma images. 

\subsection{Experimental setup}

We investigate whether we can predict the melanoma/non-melanoma labels of the images, based only on the visual assessments of the students.  This is a proof of concept to investigate whether there is any signal of crowd labels for such images - we do not propose to replace ML algorithms by crowds. However, we expect that if there is signal in the crowd labels, a ML classifier trained on image features could be improved by including crowd labels as additional input. 

We perform two experiments. For each experiment, we use 96 images (after removing four images with missing values), and perform a 10-fold cross-validation. We use a logistic classifier. This choice is based on our experience with small datasets, and was not selected to maximize performance in any particular case. Due to the class imbalance in the dataset, we use the area under the ROC curve (AUC) as the evaluation measure. 

In the first experiment, we test whether the visual assessments can be used to predict the melanoma/non-melanoma labels. We compare using all 30 features, to only features of a particular type (6 features per dataset), to all features of a particular annotator (5 features per dataset). 

In the second experiment, we test whether agreement or disagreement between annotators affect the results. For this, we use the first four distribution moments of each feature type: mean, standard deviation, skewness and kurtosis. The mean illustrates removing disagreement, while the other moments illustrate retaining disagreement only. We compare using all combined features (20 features), to using only features for each moment (5 features per dataset). 

Note that in both experiments we do not use any information directly from the image.

\section{Results}

\begin{figure}
   \centering
    \includegraphics[width=0.45\textwidth]{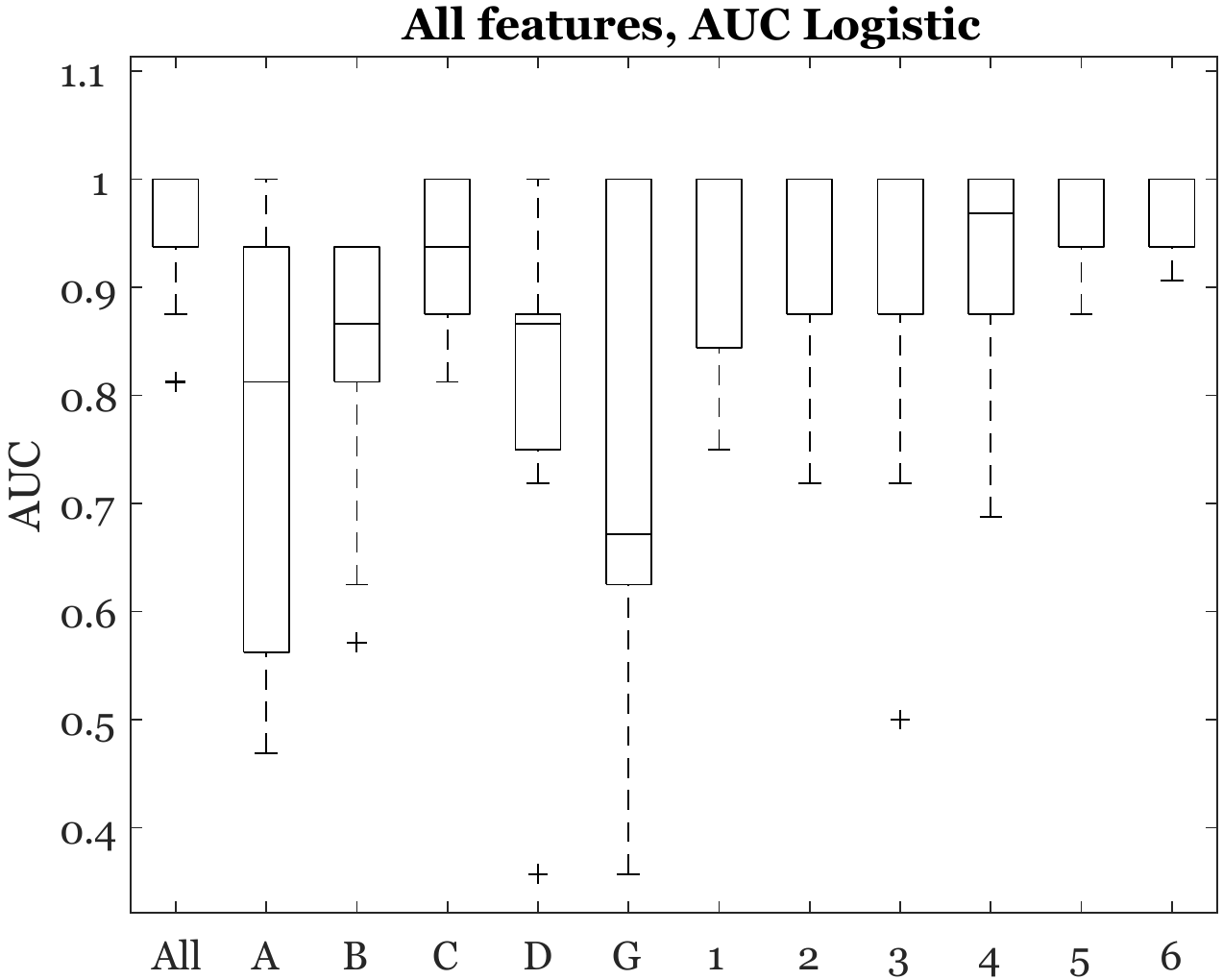}
    \includegraphics[width=0.45\textwidth]{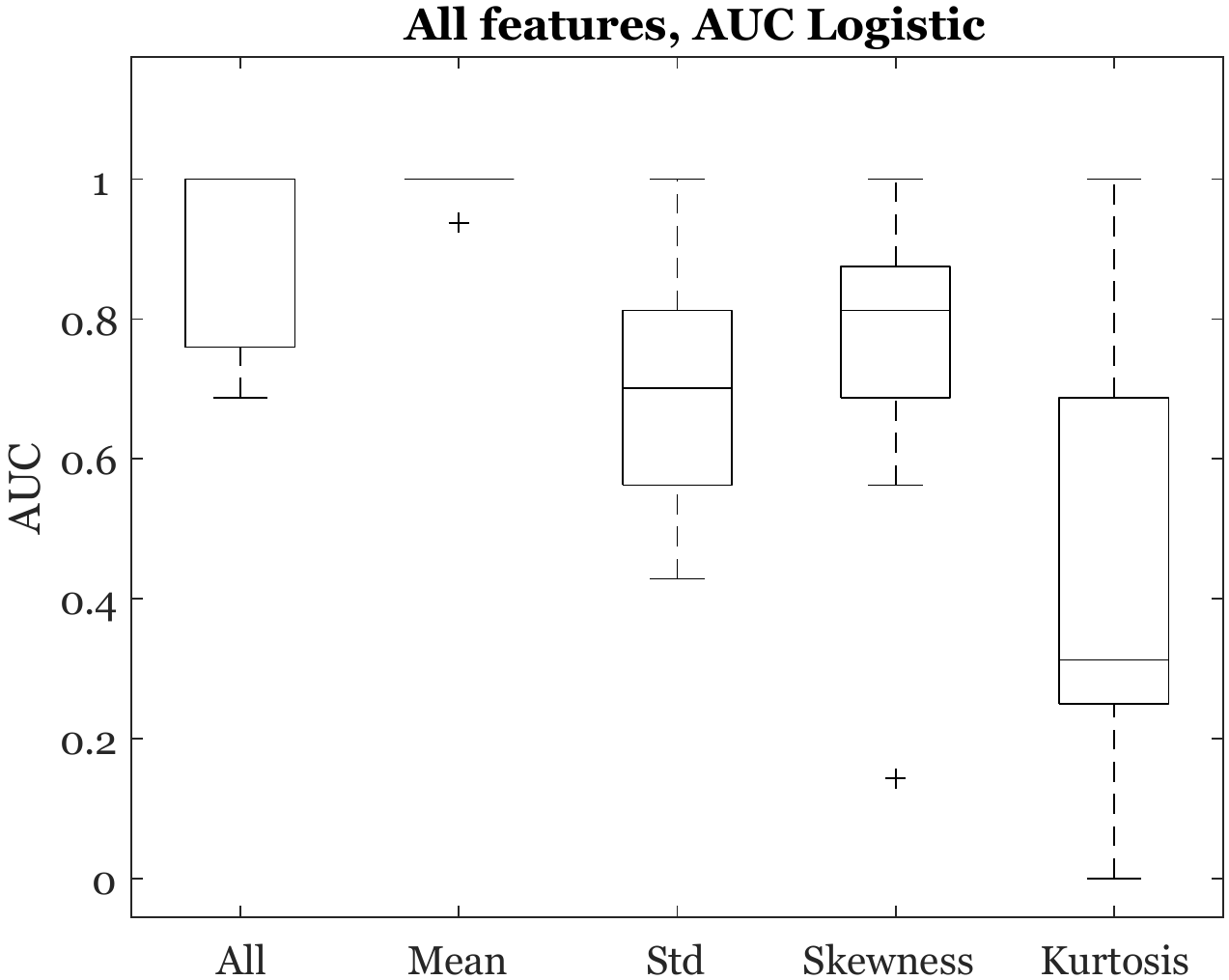}
    
    \caption{AUC performances of 10-fold cross-validation on all features. Left: only features of a particular type (\textbf{A}symmetry, \textbf{B}order, \textbf{C}olor, \textbf{D}ermoscopic structures, blue \textbf{G}low, 6 features per dataset) and only features of a particular annotator (1-6, 5 features per dataset). Right: combined features, all moments (20 features), or only moments of a particular type (mean, standard deviation, skewness, kurtosis, 5 features per dataset).}
    \label{fig:all}
\end{figure}

The results of the first experiment are shown in Fig.~\ref{fig:all} (left). Using all features leads to a very good performance (mean AUC 0.96). Using only one type of feature is worse than using all features. There are also large differences between the feature types. Color is the best feature type (mean AUC 0.93), followed by Border (mean AUC 0.82) and Dermoscopic structures (mean AUC 0.81). Other feature types are less good (mean AUC 0.78 and 0.73) but still informative. Although Glow has a reasonable average (0.73), the variability is very high, with a worse-than-random AUC in some folds. Using the features of only one annotator leads to good performance in all cases (mean AUC between 0.90 and 0.98).


The results of the second experiment are shown in Fig.~\ref{fig:all} (right). The means of the features lead to the best performance overall (mean AUC 0.99), suggesting that removing disagreement might be the best strategy. However, other moments can lead to performances that are on average better than random: mean AUC 0.71 for standard deviation and 0.73 for skewness. This suggests that there is signal in disagreement, and that it should not be removed by default. However, the variability is very high, so in some folds these features hurt, rather than help, the classifier. 

\begin{figure}
   \centering
    \includegraphics[width=0.5\textwidth]{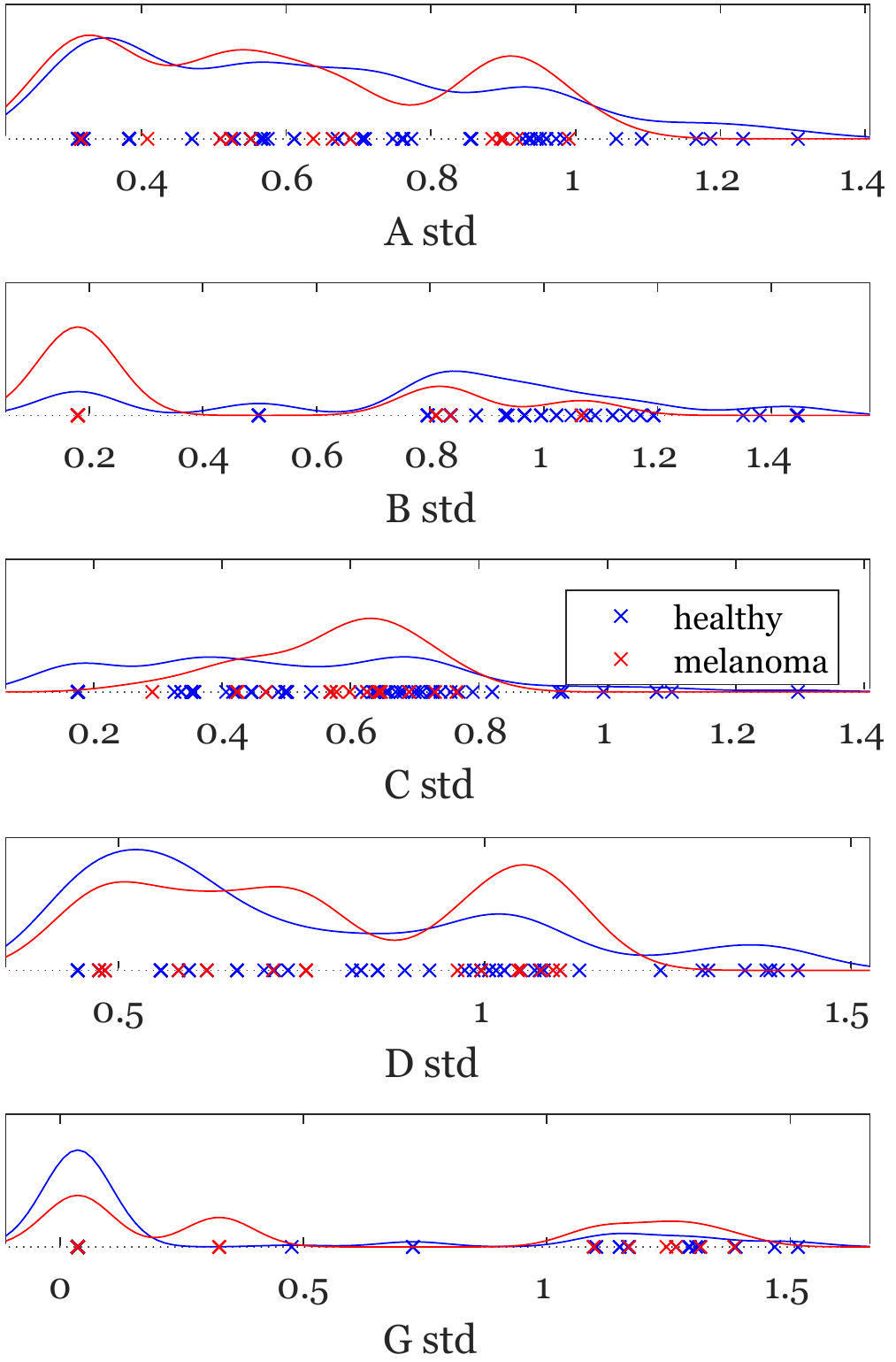}
    \caption{Distributions of the standard deviation features, capturing disagreement, for the non-melanoma and melanoma classes. Top to bottom: asymmetry, border, color, dermoscopic structures, blue glow.}
    \label{fig:plotf}
\end{figure}

To further investigate why disagreement contributes to classification, we examined the distributions of some of the moments, separately for the melanoma and non-melanoma samples. The distributions of the standard deviations (normalized to zero mean, unit variance) are shown in Fig.~\ref{fig:plotf}. There are no strong trends, but for the features A to D, more non-melanoma samples have high disagreement. This suggests that, for melanoma samples, the crowd more often agrees that the image looks abnormal.

\section{Discussion}

We used only a small dataset in these pilot experiments. Experiments with annotations collected from the other groups are needed to verify the results presented here. In particular it will be interesting to examine the influence of the number of annotators on the results, as in the other groups, each annotation was repeated by three annotators instead of six. 

We could not use typical measures for inter-observer agreement, since these would provide a scalar for any two annotators, whereas our experiment required a vector. We therefore used distribution moments, with the mean as a measure of consensus, and standard deviation, skewness and kurtosis as measures of disagreement. These are not necessarily the most suitable choices. 

An important point for future investigation is how the crowd annotations can be used to improve ML algorithms. One possibility would be to use multi-task learning to predict a vector consisting of both the expert label and (multiple) crowd annotations. For example, \cite{murthy2017center} use multi-task learning to predict the label and several visual attributes. However, the visual attributes are not provided by the crowd and consensus is already assumed. Another strategy is to first pretrain a network with the crowd annotations, and then fine-tune on the expert labels. This type of two-step strategy was successfully used with handcrafted features describing breast masses~\cite{dhungel2017deep}. However, since these features were extracted automatically, only a single feature per image was available and there was no need to address disagreement. 

\section{Conclusion}
We investigated whether disagreement between annotators could be informative. We trained a classifier to predict a melanoma/non-melanoma label from crowd assessments of visual characteristics of skin lesion images, without using the images themselves. Averaging crowd assessments to remove disagreement gave the best results, but using disagreement only gave better than random performance. 
In future work we will investigate how to integrate such crowd annotations in training an image classifier, for example via multi-task or transfer learning.


\section*{Acknowledgments}

We thank the students of the 8QA01 2017-2018 course for their participation in gathering the annotations.

\bibliographystyle{splncs}
\bibliography{refs_main}

\begin{thebibliography}{1}

\bibitem{hussein2017risk}
Hussein, S., Cao, K., Song, Q., Bagci, U.:
\newblock Risk stratification of lung nodules using {3D} {CNN}-based multi-task
  learning.
\newblock arXiv preprint arXiv:1704.08797 (2017)

\bibitem{oneil2017crowdsourcing}
O???Neil, A.Q., Murchison, J.T., van Beek, E.J., Goatman, K.A.:
\newblock Crowdsourcing labels for pathological patterns in ct lung scans: Can
  non-experts contribute expert-quality ground truth?
\newblock In: Intravascular Imaging and Computer Assisted Stenting, and
  Large-Scale Annotation of Biomedical Data and Expert Label Synthesis.
\newblock Springer (2017)  96--105

\bibitem{cheplygina2016early}
Cheplygina, V., Perez-Rovira, A., Kuo, W., Tiddens, H., de~Bruijne, M.:
\newblock Early experiences with crowdsourcing airway annotations in chest
  {CT}.
\newblock In: Large-scale Annotation of Biomedical data and Expert Label
  Synthesis (MICCAI LABELS). (2016)  209--218

\bibitem{maier2015crowdtruth}
Maier-Hein, L., Kondermann, D., Ro{\ss}, T., Mersmann, S., Heim, E.,
  Bodenstedt, S., Kenngott, H.G., Sanchez, A., Wagner, M., Preukschas, A.,
  et~al.:
\newblock Crowdtruth validation: a new paradigm for validating algorithms that
  rely on image correspondences.
\newblock International Journal of Computer Assisted Radiology and Surgery
  \textbf{10}(8) (2015)  1201--1212

\bibitem{guan2017who}
Guan, M.Y., Gulshan, V., Dai, A.M., Hinton, G.E.:
\newblock Who said what: Modeling individual labelers improves classification.
\newblock arXiv preprint arXiv:1703.08774 (2017)

\bibitem{codella2017skin}
Codella, N.C., Gutman, D., Celebi, M.E., Helba, B., Marchetti, M.A., Dusza,
  S.W., Kalloo, A., Liopyris, K., Mishra, N., Kittler, H.,  et~al.:
\newblock Skin lesion analysis toward melanoma detection: A challenge at the
  2017 {International Symposium on Biomedical Imaging (ISBI)}, hosted by the
  {International Skin Imaging Collaboration (ISIC)}.
\newblock arXiv preprint arXiv:1710.05006 (2017)

\bibitem{abbasi2004early}
Abbasi, N.R., Shaw, H.M., Rigel, D.S., Friedman, R.J., McCarthy, W.H., Osman,
  I., Kopf, A.W., Polsky, D.:
\newblock Early diagnosis of cutaneous melanoma: revisiting the abcd criteria.
\newblock Jama \textbf{292}(22) (2004)  2771--2776

\bibitem{murthy2017center}
Murthy, V., Hou, L., Samaras, D., Kurc, T.M., Saltz, J.H.:
\newblock Center-focusing multi-task {CNN} with injected features for
  classification of glioma nuclear images.
\newblock In: IEEE Winter Conference on Applications of Computer Vision (WACV),
  IEEE (2017)  834--841

\bibitem{dhungel2017deep}
Dhungel, N., Carneiro, G., Bradley, A.P.:
\newblock A deep learning approach for the analysis of masses in mammograms
  with minimal user intervention.
\newblock Medical Image Analysis \textbf{37} (2017)  114--128

\end{thebibliography}

\end{document}